\documentclass{article}

\usepackage{arxiv}
\usepackage{float}
\usepackage[utf8]{inputenc} 
\usepackage[T1]{fontenc}    
\usepackage{hyperref}       
\usepackage{url}            
\usepackage{graphicx}
\graphicspath{ {./images/} }

\title{Classification of descriptions and summary using multiple passes of statistical and natural language toolkits}

\date{} 					

\author{ \href{https://orcid.org/0000-0003-1002-1828}{\includegraphics[scale=0.06]{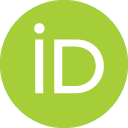}\hspace{1mm}Saumya Banthia}\\
	Synechron Innovation Lab\\
	\texttt{Saumya.Banthia@synechron.com}\\
	\And
	\href{https://orcid.org/0000-0002-9064-3362}{\includegraphics[scale=0.06]{orcid}\hspace{1mm}Anantha Sharma}\\
	Synechron Innovation Lab\\
	\texttt{Anantha.Sharma@synechron.com}\\
}

\renewcommand{\headeright}{}
\renewcommand{\undertitle}{Summary-Name-Relevance Classification}

\hypersetup{
pdftitle={Summary-Name-Relevance Classification},
pdfsubject={cs.CL},
pdfauthor={Saumya Banthia},
pdfkeywords={summary classification, definition classification, name relevance},
}

\begin{document}
\maketitle

\begin{abstract}
	This document describes a possible approach that can be used to check the relevance of a summary / definition of an entity with respect to its name. This classifier focuses on the relevancy of an entity’s name to its summary / definition, in other words, it is a name relevance check. The percentage score obtained from this approach can be used either on its own or used to supplement scores obtained from other metrics to arrive upon a final classification; at the end of the document, potential improvements have also been outlined. The dataset that this document focuses on achieving an objective score is a list of package names and their respective summaries (sourced from pypi.org \cite{pypi}).
\end{abstract}


\section{Introduction}
\label{sec:intro}
The dataset we set out to work with, contains 982 entries (982 sets of names and their summaries). The data was sourced from PyPi \cite{pypi}, which is a popular python package repository.

This dataset had the following challenges:
\begin{itemize}
	\item In many cases word abbreviations and acronyms were used in the package names, while their full forms in the summary text.
	\item Quantum of the summary text did not automatically mean a good summary.
	\item Some of the entries also have partial text or misspelling(s).
\end{itemize}

\section{About the data}
\label{sec:data}

The Python Package Index (PyPi) \cite{pypi} is a popular python programming language repository with each package having a relatively detailed summary.

The characteristics of the summary can contain description of the package along with technical keywords (which may be relevant), which are sometimes directly mentioned in the name, but in other instances, are either absent or present in abbreviated forms or as partial words. 

The quantum of text (number of word / lines of text) does not necessarily reflect the quality of the summary. Therefore, performing a broad summarization on the complete list of entities might not be the best approach. Although, it can be used to deduce some basic characteristics of the dataset. 

Therefore, we decided on an approach, which could take these inconsistencies into account and produce a viable score. 

\pagebreak
\subsection{Introduction to the tools}
The use of pre-existing tools has been done, so as not to re-invent the wheel. 

The following techniques were used:
\begin{itemize}
	\item Statistical splitting of concatenated words – wordninja \cite{wnin}: This package probabilistically splits concatenated words using NLP based on English Wikipedia unigram frequencies. 
	\item Lemmatization and Stopwords removal – NLTK \cite{nltk}
	\item Fuzzy matching – FuzzyWuzzy \cite{fuzz}: It uses Levenshtein Distance to calculate the differences between sequences.
\end{itemize}

\begin{figure}[h]
	\centering
	\includegraphics[scale=0.5]{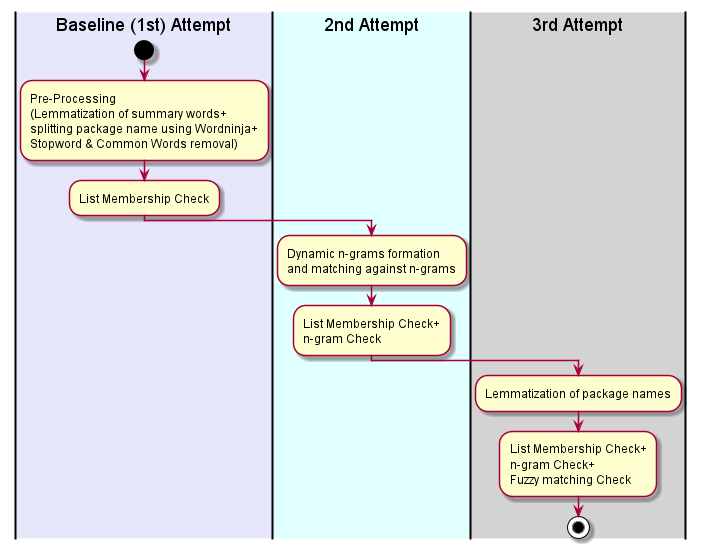}
	\caption{Incremental changes across consecutive attempts.}
	\label{fig:fig1}
\end{figure}

Similar techniques from other libraries can be used as a substitute for some of the libraries used in this case. 

Basic functionality (in the context of the use case) of each package is explained below in brief.

Since the package names were mostly concatenated into a single string (pytracks -> py + tracks, parsejson -> parse + json, etc), therefore they had to be broken down into separate words. For this purpose, wordninja \cite{wnin} was quite useful. 

Stop-Words (like “a”, “an”, “the”, etc) had be removed from the summary and package names for 2 main reasons:
\begin{itemize}
	\item To improve loss in accuracy caused due to presence / absence of these in either the package name or summary (i.e. presence in one and absence in another or vice-versa). 
	\item To improve compute times while doing membership checks.
\end{itemize}

The WordNetLemmatizer stems the words to further improve matches (for instance, “loving” which is a verb can be stemmed to “love”) by converting words into their base form (Lemmatized versions). 

FuzzyWuzzy \cite{fuzz} is used as the fuzzy matching library, which can return the closest match to a string (and its score). This can be used to improve the matching in case of partial / misspelled strings.

\section{The Approach}
\label{sec:approach}
Several attempts were made, with each attempt building on top of the previous one. We will also go over the cons of each attempt (until the last one) to grasp a better understanding of the changes that were introduced with each new attempt.

\subsection{1\textsuperscript{st} Attempt (Baseline)}

To establish a baseline, a vanilla membership check was performed (only checking if a specific word from the package name occurs in the summary text). 

The transformations applied at this stage were the removal of common words (in our case there were 2 [“py”, “python”], which we determined by obtaining summary statistics on the 2 columns - namely, package name and summary), removal of stop words, lemmatization of the words in the summary and splitting the concatenated package names using wordninja \cite{wnin}.

\begin{figure}[h]
	\centering
	\includegraphics[scale=0.75]{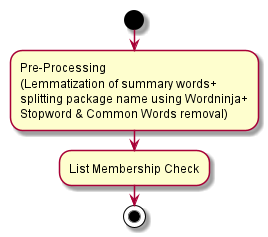}
	\caption{Baseline attempt pipeline.}
	\label{fig:fig2}
\end{figure}

\begin{figure}[h]
	\centering
	\includegraphics[scale=0.75]{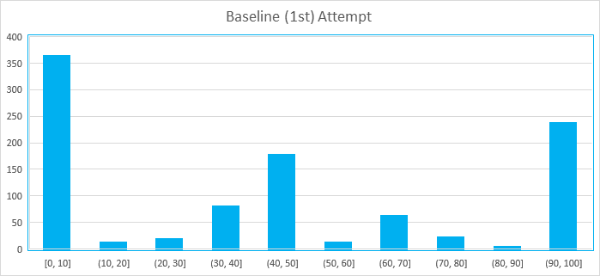}
	\caption{Scores from Baseline Attempt}
	\label{fig:fig3}
\end{figure}

This produced 362 entries marked with a score of 0 (out of 100), and 236 entries scoring 100. This meant there was a huge scope for improvement. [Figure \ref{fig:fig3}] shows a better summary, from the baseline attempt.

\subsubsection{Cons}
The Baseline attempt had the following cons:

\begin{itemize}
	\item "wordninja" \cite{wnin} library produced incorrectly split words in some instances, which caused bad matches. This remained a constant, due to lack of a better statistical string splitting tool.
	\item In many cases acronyms were used in the package names, while their full forms in the summary text.
\end{itemize}
\raggedbottom

\subsection{2\textsuperscript{nd} Attempt}

Addition of a dynamic n-gram generating function improved the score in some cases.

This addition reduced the number of zero scoring entries from 362 to 334 and brought up the number of entries scoring a 100 from 236 to 251.

\begin{figure}[H]
	\centering
	\includegraphics[scale=0.75]{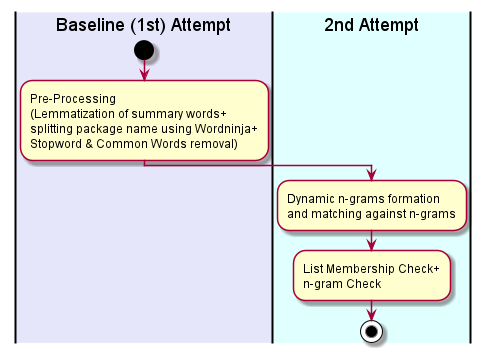}
	\caption{2\textsuperscript{nd} attempt pipeline.}
	\label{fig:fig4}
\end{figure}

\begin{figure}[H]
	\centering
	\includegraphics[scale=0.75]{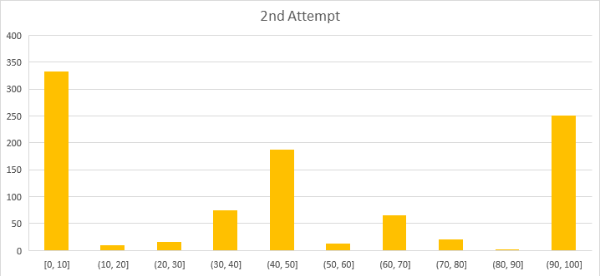}
	\caption{Scores from 2\textsuperscript{nd} Attempt}
	\label{fig:fig5}
\end{figure}

These were not huge improvements, therefore there still was scope for improvement. [Figure \ref{fig:fig5}] shows a better summary, from the 2\textsuperscript{nd} attempt.

\subsubsection{Cons}
The 2\textsuperscript{nd} attempt had the following cons:

\begin{itemize}
	\item Many words were not getting matched due to lemmatization inconsistencies (for instance “logging” has a lemmatized version “log”).
	\item Some words had spelling error(s), while others were partial words, which was leading to elements not getting matched.
\end{itemize}
\pagebreak

\subsection{3\textsuperscript{rd} Attempt}

The 3\textsuperscript{rd} attempt added a Lemmatization check and a Fuzzy matching check with a threshold greater than 25\% (anything below 25\% would not be considered a good fuzzy match, hence would not affect the final score).

\begin{figure}[H]
	\centering
	\includegraphics[scale=0.75]{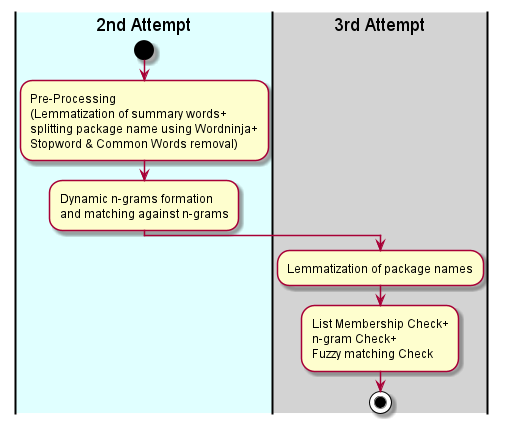}
	\caption{3\textsuperscript{rd} attempt pipeline.}
	\label{fig:fig6}
\end{figure}

\begin{figure}[h]
	\centering
	\includegraphics[scale=0.75]{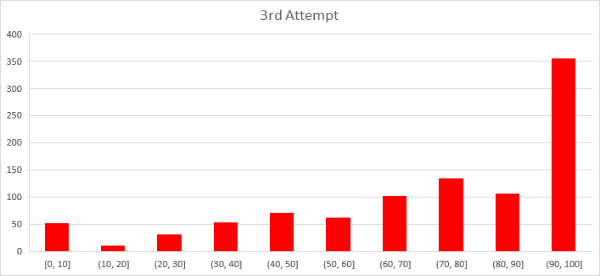}
	\caption{Scores from 3\textsuperscript{rd} Attempt}
	\label{fig:fig7}
\end{figure}

These changes introduced by far the most drastic delta in scores. The total number of zero scoring entries got down to 50 (from 334) although there was not a huge improvement in the number of 100 scoring entries, which increased to 256 (from 251). [Figure \ref{fig:fig7}] shows a better summary, from the 3\textsuperscript{rd} attempt.

\pagebreak

\subsection{Comparing the 3 Attempts}

The bottom 50\% scores (scores less than 50) accounted for almost 50\% of the entries, which got shaved down by more than half its original amount to about 21\%.

\begin{figure}[h]
	\centering
	\includegraphics[scale=0.75]{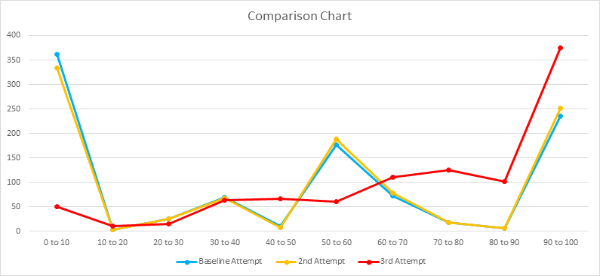}
	\caption{Comparison of scores among the 3 attempts}
	\label{fig:fig8}
\end{figure}

\pagebreak
To validate the scores, a manual review of package name-summary pairs was performed. A set of 101 entries were manually scored. The labels used are described below:
\begin{itemize}
	\item (0: Unknown -> Range: 0-25)
	\item (1: Poor -> Range: 25-50)
	\item (2: Good -> Range: 50-75)
	\item (3: Perfect -> Range: 75-100)
\end{itemize}

The following rules were kept in mind while scoring
\begin{itemize}
	\item If the name of package occurs in the summary, a perfect or good score was given. 
	\item A “0” was given only if the summary is either too generic, or not present.
	\item If the package has a non-English vocabulary (but contains elements from the package name) score it either perfect or good.
\end{itemize}

Each package name-summary pair was allotted 2 scores (one primary, one secondary), if the predicted score is equal to the primary score, a validation score of 1 was given, if it was equal to the secondary score, then 0.5, else, 0 (i.e. if the predicted score was different from both primary and secondary scores).

\begin{figure}[h]
	\centering
	\includegraphics[scale=0.75]{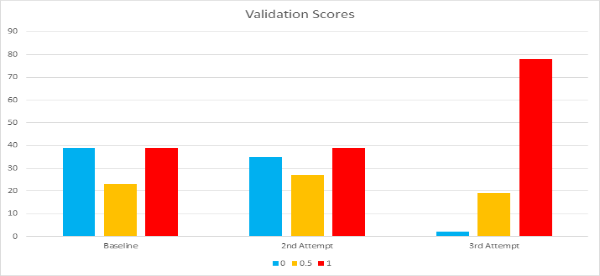}
	\caption{Comparison of validation scores among the 3 attempts}
	\label{fig:fig9}
\end{figure}

Through the manual review, the 3\textsuperscript{rd} attempt stood out, in comparison to the baseline and 2\textsuperscript{nd} attempt scores. [Figure \ref{fig:fig9}] shows the validation scores across the 3 attempts.

A large chunk of the scores improved in the 3\textsuperscript{rd} attempt, this was partly due to the addition of a Lemmatization check added to the package names (earlier only summary words were getting lemmatized) but, can be majorly attributed to fuzzy matching check, as it helped deal with cases of partial matches and misspelled words.

\pagebreak
\section{Conclusion}
\label{sec:conclusion}
The final attempt mentioned in this document seems to produce the most accurate results. This approach can not only be used directly, but also to supplement results generated from other approaches for a better final classification (which is also a recommended way to use it).

All of this is considering the following factor(s):
\begin{itemize}
	\item The dataset had not much contributed to it in terms of pre-processing / cleaning.
\end{itemize}

These were only a small proportion of the approaches that could be applied to solve a problem like this. There is scope for improvement in these approaches, some of which have been outlined in the next section.

\section{Recommendations}
\label{sec:recommendations}
\begin{itemize}
	\item Use better data pre-processing pipelines to improve accuracy of results obtained.
	\item "wordninja" \cite{wnin} was trained on data from Wikipedia. The library has the capability to integrate custom language models which could be more contextually relevant. This in turn can be used to improve splitting accuracy and therefore reduce matching complexity.
	\item Algorithms to deduce data types, as well as intent (in case of input fields and their definitions) can be used to supplement scores obtained from the current approach.
	\item Using tf-idf approach to find key words to focus on might further improve accuracy.
\end{itemize}

\bibliographystyle{unsrt}
\bibliography{references}

\end{document}